\def\arxivversion{}
\documentclass[letterpaper, 10pt, conference]{ieeeconf}
\IEEEoverridecommandlockouts
\input{preamble}

\title{\LARGE \bf \textsc{Specola}: A Tool for Specification Driven Robot Evaluation}

\ifdefined\arxivversion
\AddToHookNext{shipout/foreground}{\put(0.5\paperwidth,-0.45in){\makebox[0pt]{%
\parbox{0.8\paperwidth}{\centering\footnotesize This work has been submitted
to the IEEE for possible publication. Copyright may be transferred without
notice, after which this version may no longer be accessible.}}}}
\author{Immanuel Ampomah Mensah and Hadas Kress-Gazit, Cornell University}
\else
\author{Omitted for anonymous review}%
\fi

\date{}

\begin{document}
\bstctlcite{IEEEexample:BSTcontrol}

\maketitle

\begin{abstract}
Simulation benchmarks are a useful way to measure progress in robotics; typically they provide success rate as the dominant measure of performance. However, success rate reports only whether a rollout completed the task, hiding by what margin it succeeded, how sensitive the success is to goal parameter changes, requirements it may have violated, and whether success persisted. Signal Temporal Logic (STL) provides a formal language for capturing specifications; it can be used to define requirements, and measure \textit{how close} a robot behavior is to satisfying or violating them.  %
Using STL monitors in a simulator, however, requires exposing simulator state and mapping it to specification-relevant signals. %
We present \toolname{}, a tool for evaluating robot behaviors in simulation with respect to STL specifications. \toolname{} provides an interface for writing specifications over simulator signals, and creates and manages the relevant monitors while a policy executes. We provide a Python API and command-line interface for integrating monitoring into existing policy execution loops. We provide an adapter interface for connecting new simulators, and provide adapters for nine simulation environments and benchmarks, including LIBERO, RoboCasa, Isaac Lab, and RoboDojo. We provide case studies that illustrate the nuanced evaluation that \toolname{} enables. %
Code and adapters are available at \repourl{}.

\end{abstract}

\section{Introduction}

The recent rapid advances in generative AI, foundation models, and agentic AI, result in a fast iteration cycle for robot policies that leverage these technologies to create robots capable of complex tasks. An enabler for rapid progress in robot learning, and robotics more generally, is the use of simulations and simulation benchmarks~\cite{yu2019metaworld,liu2023libero,nasiriany2024robocasa}.

The most common metric used in the community to assess progress and relative performance of robot policies is through the use of \textit{success rate}--the percentage of robot executions or rollouts that result in task success, as determined by the benchmark. While
success rate is useful for comparing outcomes, it reduces our understanding of \textit{how} a policy performs. Two unsuccessful rollouts may differ in how close they come to satisfying the task or in which parts of it they complete, and a successful rollout might be on the cusp of failing. Furthermore, even the division between success and failure depends on thresholds defined by a benchmark, for example how close an object is to a target pose; policies successfully executing a task in different ways may exhibit different success rates, since success depends on the chosen threshold, thus providing a misleading signal to the community regarding progress in the field. %

Signal Temporal Logic (STL)\cite{maler2004monitoring}, is a formal specification language that provides a rigorous foundation for expressing requirements over continuous signals--here robot and environment state. In addition to Boolean semantics (True/False), it admits a qualitative semantics; it can be used to calculate, for a given rollout, a signed \textit{robustness score} that indicates not only if the specification holds, but \textit{how close} the rollout is to satisfying or violating the specification. 

Tracking a rollout's robustness score with respect to a specification is done through STL \textit{monitors}. These monitors use continuous signals and an STL formula to produce the score. To run an STL monitor in simulation requires exposing the simulator state (robot and environment state) and mapping it to the formula-relevant signals. %

\noindent\textbf{Contributions:} We present \textbf{\toolname{}}, a tool for evaluating robot behavior in simulation using STL specifications and monitors it automatically creates and deploys to track the robustness score. \toolname{} (i) automatically detects all the signals that can be monitored, (ii) includes a specification writing interface that facilitates writing specifications, (iii) provides a Python API and command-line interface that integrate monitoring into existing policy execution loops and report robustness scores during a rollout, and (iv) in order to make \toolname{} interface with any simulator, we have designed an adapter interface. We provide adapters for RoboCasa~\cite{nasiriany2024robocasa}, LIBERO~\cite{liu2023libero}, Meta-World~\cite{yu2019metaworld}, dm\_control~\cite{tunyasuvunakool2020}, MA-MuJoCo~\cite{peng2021facmac}, PyBullet~\cite{coumans2021}, Isaac Lab~\cite{mittal2025isaaclab}, RoboDojo~\cite{chen2026robodojo}, and CARLA~\cite{dosovitskiy2017carla}. We also provide instructions on how to write an adapter for other simulators. In the following we show case studies for the insight \toolname{} can provide regarding policy performance.

\section{Related Work}
\subsection{Evaluation of Learned Robotic Policies}
Kress-Gazit \emph{et al.} criticize reliance on success rate as the sole criterion for evaluating robot policies because it provides little information about execution quality or failure modes~\cite{kressgazit2024empirical}. One way to add detail is to measure partial completion. BEHAVIOR measures the fraction of satisfied goal predicates~\cite{srivastava2021behavior}, whereas MetaFine evaluates completion of constituent skill stages against their acceptance criteria~\cite{xu2026metafine}. These approaches expose progress that an overall binary outcome can hide. Snyder \emph{et al.} address the statistical comparison of policies through a sequential framework for bounded evaluation scores~\cite{snyder2026beyondbinary}. Their experiments show that partial-credit scores can reduce the number of evaluation trials relative to binary outcomes.

Other evaluations measure compliance with safety requirements alongside task completion. SafeVLA-Bench provides the closest comparison to our evaluation, formalizing safety requirements as STL formulas and reporting a success-safety gap on LIBERO and RoboCasa through post-hoc analysis~\cite{fan2026safevla}. %
LIBERO-Safety reports a collision rate from hand-coded contact checks, without a specification language~\cite{cui2026liberosafety}.

To check a specification, an evaluator must connect its predicates and entities to simulator state. MANGO generates pass/fail checks %
from hand-supplied predicate and object vocabularies and localizes failures to a step index~\cite{valle2026mango}. SafeManip checks temporal logic safety templates (LTLf~\cite{clarke2018model}) on recorded RoboCasa rollouts and reports whether each template was satisfied or violated; its predicates are hand-written inside a modified copy of the simulator~\cite{huang2026safemanip}. SENTINEL checks temporal logic (LTL and CTL) specifications post hoc; its authors leave online checking and STL as future work~\cite{zhan2026sentinel}. LabGuard matches the entity names in a rule against a hand-maintained registry~\cite{yang2026labguard}. %
These tools require predicates and entities in advance, whereas \toolname{} automatically discovers entities and monitorable signals from the connected simulator.

More recently, XPolicyLab standardized the interface between policies and evaluation environments, so that a policy is integrated once and evaluated in many environments~\cite{community2026xpolicylab}. \toolname{} addresses a different part of the evaluation, the connection between specifications and simulator state.

\subsection{STL Monitoring Tools and Semantics}
Online STL monitoring updates robustness scores as new signal samples arrive~\cite{deshmukh2017robust}. %
Tools such as RTAMT~\cite{nickovic2020rtamt}, Breach~\cite{donze2010breach}, and PerceMon~\cite{balakrishnan2021percemon} operate on signal traces. To apply these tools to a simulator, an evaluator must map simulator state to the signals in the specification. We developed \toolname{} to provide this integration through reusable adapters and a discovery interface.

\section{Preliminaries}

\label{sec:prelim}
We monitor the past-time fragment of STL~\cite{maler2004monitoring,donze2010robust} over discrete-time signals. A trace is a signal $x : \mathbb{N} \to \mathbb{R}^n$ where the signal is sampled once per timestep.

\noindent \textbf{Syntax:} Given atomic predicates of the form $\mu \equiv f(x) \geq 0$ with $f : \mathbb{R}^n \to \mathbb{R}$, past-time STL formulas are defined as: %
\begin{equation*}
\varphi ::= \mu \mid \neg\varphi \mid \varphi_1 \wedge \varphi_2 \mid \Gop{[a,b]}\,\varphi \mid \Fop{[a,b]}\,\varphi \mid \varphi_1\, \Sop{[a,b]}\, \varphi_2,
\end{equation*}
where $0 \leq a \leq b$ are time bounds in steps and each temporal operator looks back over the trailing window $[t-b,\,t-a]$. \emph{Historically}, $\Gop{[a,b]}\,\varphi$, requires $\varphi$ to hold at every step of the window. \emph{Once}, $\Fop{[a,b]}\,\varphi$, requires $\varphi$ to hold at some step of the window. \emph{Since}, $\varphi_1\, \Sop{[a,b]}\, \varphi_2$, requires $\varphi_2$ to hold at some step of the window and $\varphi_1$ to hold at every step after it, up to the current one.

\noindent \textbf{Boolean Semantics:} The satisfaction relation $(x,t) \models \varphi$ is defined inductively by
\begin{align*}
&(x,t) \models \mu \;\text{iff}\; f(x(t)) \geq 0,\\
&(x,t) \models \neg\varphi \;\text{iff}\; (x,t) \not\models \varphi,\\
&(x,t) \models \varphi_1 \wedge \varphi_2 \;\text{iff}\; (x,t) \models \varphi_1 \text{ and } (x,t) \models \varphi_2,\\
&(x,t) \models \Gop{[a,b]}\varphi \;\text{iff}\; \forall t' \in [t-b,\,t-a] : (x,t') \models \varphi,\\
&(x,t) \models \Fop{[a,b]}\varphi \;\text{iff}\; \exists t' \in [t-b,\,t-a] : (x,t') \models \varphi,\\
&(x,t) \models \varphi_1 \Sop{[a,b]} \varphi_2 \;\text{iff}\; \begin{aligned}[t] &\exists t' \in [t-b,\,t-a] : (x,t') \models \varphi_2 \\ &\text{and } \forall t'' \in (t',\,t] : (x,t'') \models \varphi_1, \end{aligned}
\end{align*}
with $t', t'' \in \mathbb{N}$.

\noindent \textbf{Quantitative Semantics:} Denoting $\rob_\varphi(t)$ as the \textit{robustness score} of $\varphi$ over $x$ at step $t$, the quantitative semantics is defined inductively by
\begin{align*}
&\rob_\mu(t) = f(x(t)),\\
&\rob_{\neg\varphi}(t) = -\rob_\varphi(t),\\
&\rob_{\varphi_1 \wedge \varphi_2}(t) = \min\bigl(\rob_{\varphi_1}(t),\, \rob_{\varphi_2}(t)\bigr),\\
&\rob_{\Gop{[a,b]}\varphi}(t) = \min_{t' \in [t-b,\,t-a]} \rob_\varphi(t'),\\
&\rob_{\Fop{[a,b]}\varphi}(t) = \max_{t' \in [t-b,\,t-a]} \rob_\varphi(t'),\\
&\rob_{\varphi_1 \Sop{[a,b]} \varphi_2}(t) = \max_{t' \in [t-b,\,t-a]} \min\Bigl(\rob_{\varphi_2}(t'), \min_{t'' \in (t',\,t]} \rob_{\varphi_1}(t'')\Bigr).
\end{align*}
The robustness score sign corresponds to the Boolean satisfaction: $\rob_\varphi(t) > 0$ iff $(x,t) \models \varphi$ and $\rob_\varphi(t) < 0$ iff $(x,t) \not\models \varphi$. The robustness score's absolute value indicates \textit{how close} a signal is to satisfying (if $\rob_\varphi(t) < 0$) or violating (if $\rob_\varphi(t) > 0$) a specification. %

\section{\toolname{} workflow}
\label{sec:system}
\begin{figure*}[t]
\centering
\includegraphics[width=\textwidth]{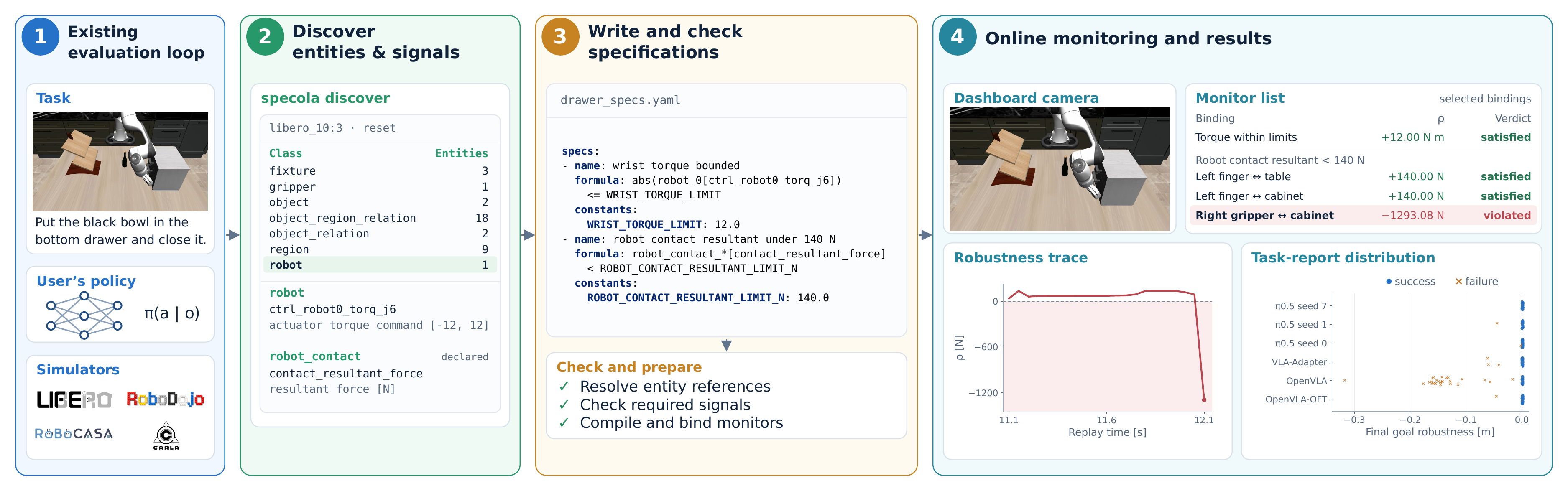}
\caption{Workflow overview. From an existing evaluation loop (1), we discover the simulator's signals (2), write and check specifications over them (3), and monitor them while the policy runs (4).}
\label{fig:arch}
\end{figure*}

The \toolname{} workflow (Fig.~\ref{fig:arch}) consists of discovering the simulator's entities and signals
(Section~\ref{sec:discovery}), writing specifications over them
(Section~\ref{sec:authoring}), and monitoring them within the policy evaluation
loop (Section~\ref{sec:running}). Table~\ref{tab:cli} summarizes the command-line interface. Connecting another simulator requires an adapter (Section~\ref{sec:adapter}); We provide adapters for RoboCasa~\cite{nasiriany2024robocasa}, LIBERO~\cite{liu2023libero}, Meta-World~\cite{yu2019metaworld}, dm\_control~\cite{tunyasuvunakool2020}, MA-MuJoCo~\cite{peng2021facmac}, PyBullet~\cite{coumans2021}, Isaac Lab~\cite{mittal2025isaaclab}, RoboDojo~\cite{chen2026robodojo}, and CARLA~\cite{dosovitskiy2017carla}. %
 
We illustrate the workflow using the following task: %
\begin{example}
\label{exm:mot}
Consider a policy performing the
LIBERO task ``put the black bowl in the
bottom drawer of the cabinet and close it''~\cite{liu2023libero}. To examine its behavior, we monitor several signals, including bowl position, drawer pose, gripper state, and force at the end effector. 
\end{example}

\subsection{Discovering signals}
\label{sec:discovery}

To write specifications, we need to know the simulator's signals. The interactive
\texttt{discover} display lists entities by class and role, with their signals,
units, and reference frames where available.

An \emph{entity} represents an identifiable scene element, such as the
bottom-drawer region. A \emph{class} is a category of entities, such as \texttt{robot}, \texttt{object}, or \texttt{region}. Within each class, the \texttt{discover} interface lists entities, grouped
by their task roles when the adapter supplies this information. In LIBERO, for
example, some objects have the role \texttt{target}
while others have the role \texttt{distractor}.

\begin{figure}[!htb]
\centering
\includegraphics[width=\columnwidth]{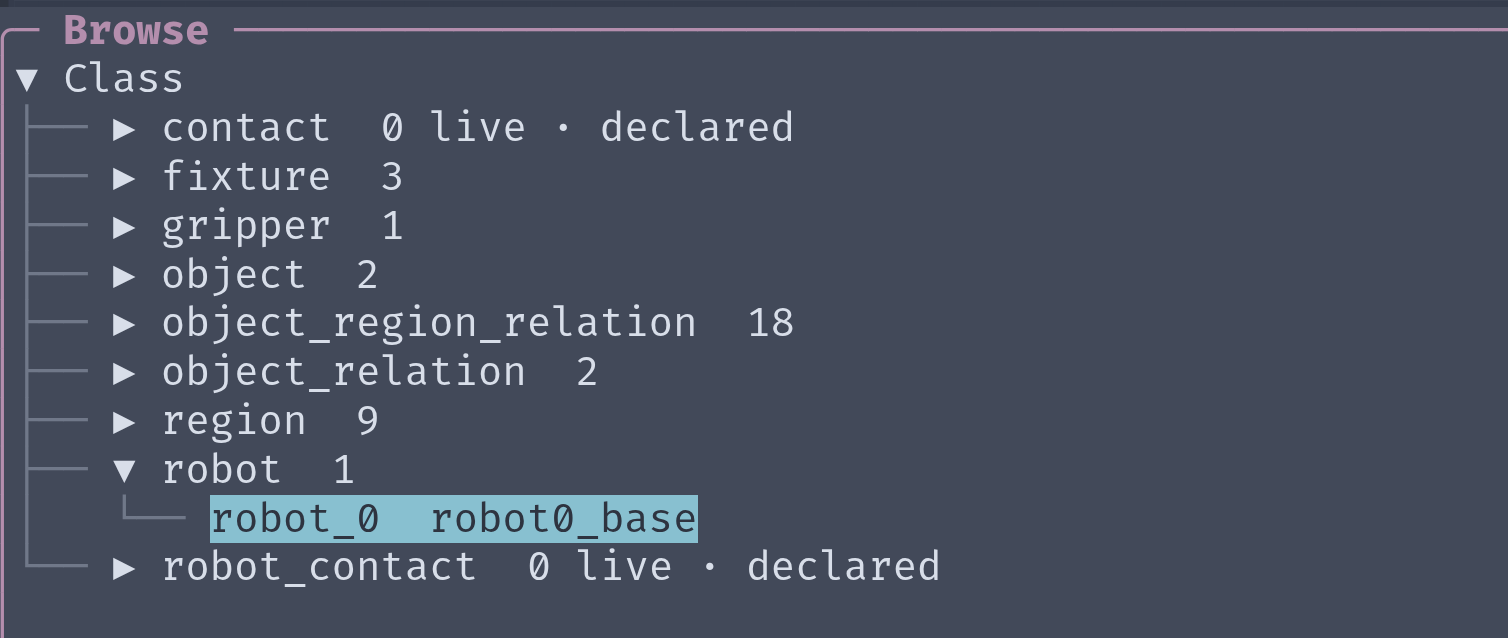}\par
{\small (a) Entities organized by class.}
\par\smallskip
\includegraphics[width=\columnwidth]{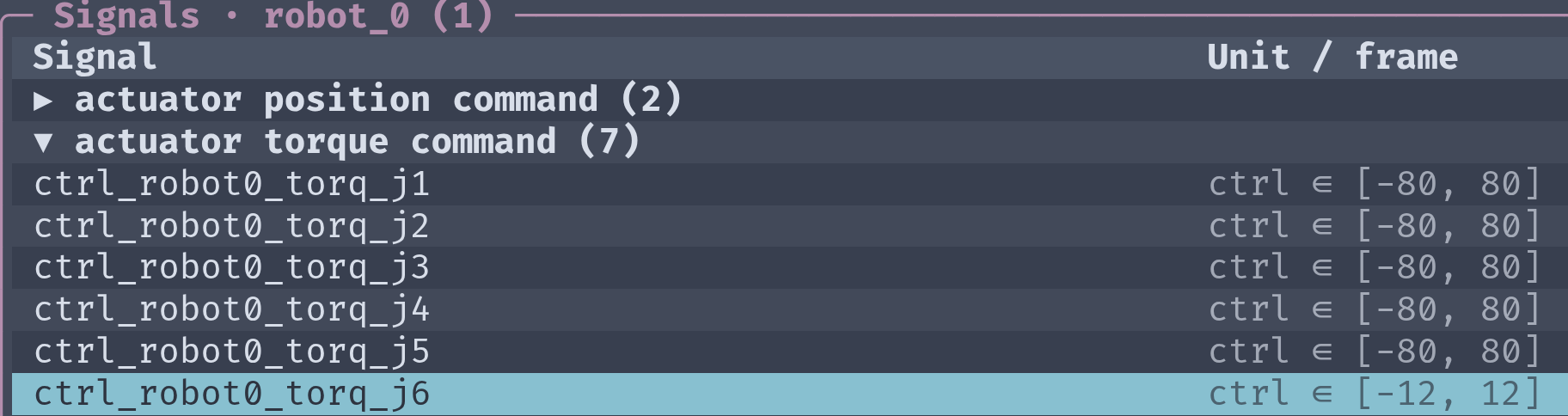}\par
{\small (b) Signals of the selected robot.}
\par\smallskip
\includegraphics[width=\columnwidth]{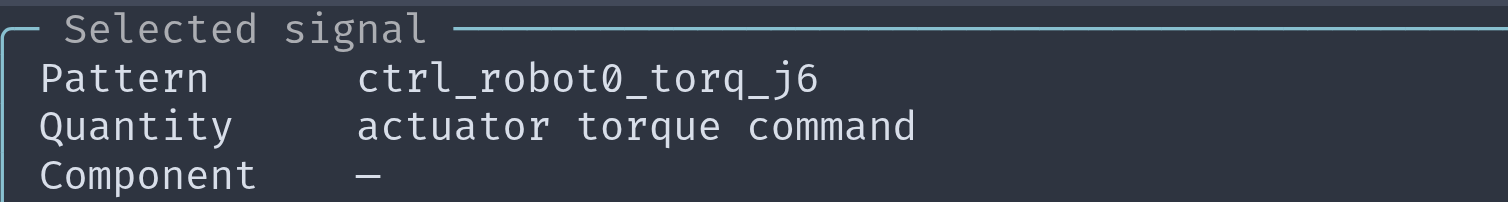}\par
{\small (c) Metadata for the selected signal.}\par
\vspace{4pt}
\caption{Discovery for the drawer task: the robot within its class (a), the properties of the robot with the wrist torque command selected (b), and the metadata of the selected signal (c).} %
\label{fig:discovery-drawer}
\end{figure}

Each entity exposes named scalar \emph{properties}, whose values over
time form the signals we monitor. 
For vector quantities, the panel also identifies the selected component
and reference frame when this metadata is available. Pairwise classes such as \texttt{robot\_contact} may list no entities at the start of a
rollout (Fig.~\ref{fig:discovery-drawer}(a)); their entities appear as interactions occur.

For the drawer task in Example~\ref{exm:mot}, we run:
\begin{lstlisting}[style=shellcommand]
specola discover --sim libero \
  --task libero_10:3 --ui textual
\end{lstlisting}
Here, \inlinecode{-{}-sim} selects the simulator adapter, \inlinecode{-{}-task}
identifies the LIBERO suite and task index, and \inlinecode{-{}-ui textual}
requests the terminal explorer. The \texttt{discover} interface lists \texttt{robot\_0} under \texttt{robot}
(Fig.~\ref{fig:discovery-drawer}(a)). Selecting it shows its properties by quantity
(Fig.~\ref{fig:discovery-drawer}(b)).
Under \emph{actuator torque command} we find
\texttt{ctrl\_robot0\_torq\_j6}, the torque command for one of the two
wrist joints. Selecting this property opens its metadata panel
(Fig.~\ref{fig:discovery-drawer}(c)).

\subsection{Writing specifications}
\label{sec:authoring}

With the signals discovered, we can write specifications directly in YAML
or compose them interactively in \toolname{}. Table~\ref{tab:stl-syntax} shows how
the operators of the grammar in Section~\ref{sec:prelim} are written in \toolname{}.

\begin{table}[!htb]
\caption{STL operators and how they are written in \toolname{}.}
\label{tab:stl-syntax}
\centering
\footnotesize
\setlength{\tabcolsep}{4pt}
\begin{tabular}{@{}lll@{}}
\toprule
\textbf{STL} & \textbf{Meaning} & \textbf{\toolname{} syntax} \\
\midrule
$\neg$ & not & \texttt{not}, \texttt{!} \\
$\wedge $ & and & \texttt{and}, \texttt{\&\&} \\
$\vee $ & or & \texttt{or}, \texttt{||} \\
$ \rightarrow $ & implies & \texttt{->} \\
$\Gop{[a,b]}$ & historically & \texttt{historically[a,b]}, \texttt{G[a,b]} \\
$\Fop{[a,b]}$ & once & \texttt{once[a,b]}, \texttt{F[a,b]} \\
$\Sop{[a,b]}$ & since & \texttt{since[a,b]} \\
\bottomrule
\end{tabular}
\end{table}

\begin{figure}[!htb]
\centering
\includegraphics[width=\columnwidth]{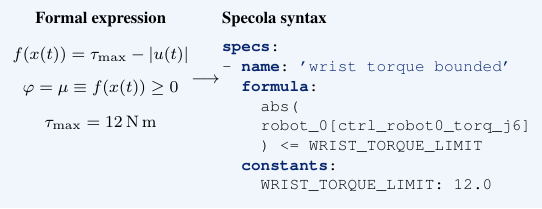}
\caption{From a formal expression to \toolname{} syntax. %
}
\label{fig:predicate-yaml}
\end{figure}

For example, let $u(t)$ denote a wrist actuator's torque command and
$\tau_{\max}$ its permitted magnitude. We define
the predicate function $f(x(t)) = \tau_{\max} - |u(t)|$.
The STL formula is then
$\varphi = \mu \equiv (\tau_{\max} - |u(t)| \geq 0)$,
or equivalently $|u(t)| \leq \tau_{\max}$.
Fig.~\ref{fig:predicate-yaml} expresses this formula in YAML using supported
arithmetic and comparison symbols. The selector identifies \texttt{robot\_0}
and the wrist-command property selected in Fig.~\ref{fig:discovery-drawer}(c).

Each entry in the YAML \texttt{specs} list has a \texttt{name}, a
\texttt{formula}, and optional \texttt{constants}. Here,
\texttt{WRIST\_TORQUE\_LIMIT} sets the bound to 12~N\,m.

To illustrate a temporal operator, we extend the wrist requirement to both wrist commands over a one-second window (Listing~\ref{lst:spec}, lines~14--16).

To facilitate writing specifications over many signals, we introduce quantification: %
A \emph{selector}, \texttt{class\_entity[properties]}, references signals
by class, entity, and property. A number selects one entity
(\texttt{robot\_0}); \texttt{*} selects every entity of the class with the
requested properties (\texttt{region\_*}). Brackets contain a property
name, a set of names (\texttt{\{a,b\}}), or \texttt{*} for all properties.
An optional \texttt{@} followed by a quantity name from \texttt{discover}
keeps only properties of that quantity.

For the task in Example~\ref{exm:mot}, we are interested in monitoring
the following requirements:
\begin{itemize}
    \item The bottom drawer is closed.
    \item Every contact force on the robot stays below 140\,N.
    \item Both wrist torque commands stay within 12\,N\,m over the last
    second.
    \item All actuator torque commands stay within 80\,N\,m over the last
    second.
\end{itemize}

Listing~\ref{lst:spec} shows the specification file for these requirements, with one entry for each.
The drawer requirement (lines~2--7) matches only the bottom-drawer
region, so it becomes one monitor. The contact requirement (lines~8--12)
becomes one monitor for each contact between a robot body and another body. %
Each torque requirement (lines~13--24) compares the largest selected command magnitude with its limit, so it becomes one monitor for the robot.

Listing~\ref{lst:spec}, lines~4 and~9, select entities with \texttt{*}, and
\texttt{robot\_0[* @ actuator torque command]} selects the
robot's seven actuator torque commands, without listing
their property names (lines~20--21). Lines~14--16 select both wrist torque commands with one selector.
Additional selector forms are described in
the tool documentation.

\begin{codelisting}[!htb]
\includegraphics[width=\columnwidth]{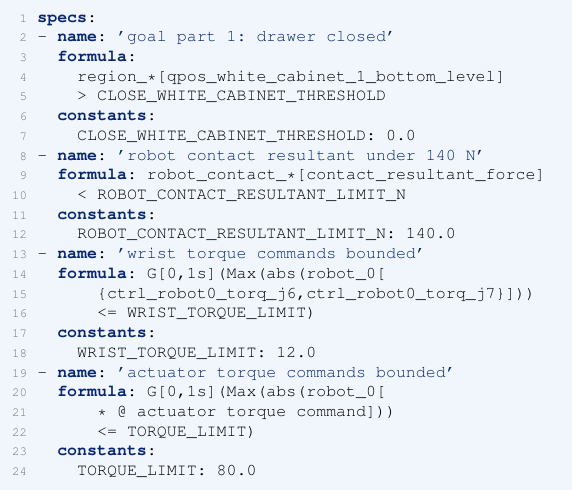}
\caption{Specification file for Example~\ref{exm:mot} showing the use of selectors.} %

\label{lst:spec}
\label{lst:final-specs}
\end{codelisting}

Specifications can also be composed interactively with
\texttt{write-specs}, which validates each formula against the grammar
and the connected scene before saving the collection to YAML (here
\texttt{drawer\_specs.yaml}, named by \inlinecode{-{}-output}). A manually written specification file
can be checked the same way:
\begin{lstlisting}[style=shellcommand]
specola check --sim libero --task libero_10:3 \
  drawer_specs.yaml
\end{lstlisting}
Before execution, this check resolves entity references and signal names
against the connected scene.

\subsection{Running Monitors}
\label{sec:running}
With our specifications written and checked, we can monitor behavior online or evaluate saved rollouts. 

\noindent \textbf{Within a policy execution loop:}  \toolname{}'s Python API adds monitoring to existing policy evaluation code with an import and four calls,
shown in bold blue in the simplified LIBERO/OpenVLA-OFT loop of Listing~\ref{lst:loop}.

We first import \inlinecode{Monitor} (line~4) from the \toolname{} package, then instantiate it with
the specification file and a directory for the results (lines~11--12).
Lines~5--10 select the LIBERO task and build its environment and initial states.
After environment setup, we call \inlinecode{monitor.watch} to attach a
monitoring session (lines~15--17), passing the benchmark task object as
\inlinecode{task} and the control-step interval in seconds as \inlinecode{dt},
computed from the environment's control frequency. \toolname{} identifies the adapter, discovers
the scene, and compiles and binds the specifications to the available
signals.
A \emph{binding} associates a specification with the scene signals used
to evaluate it. %

We then call \inlinecode{session.step()} to evaluate the initial state
(line~18). The boxed part of the loop represents policy inference and action
selection; it is followed by another call to
\inlinecode{session.step()} (line~23). This gives the monitor one observation per executed action, even when the policy returns action chunks. The returned result
contains robustness scores and truth values for the specifications. %

At rollout end, \inlinecode{session.end} closes the monitoring episode and
saves the benchmark's success flag with the results (line~25).
\begin{codelisting}[!htb]
\includegraphics[width=\columnwidth]{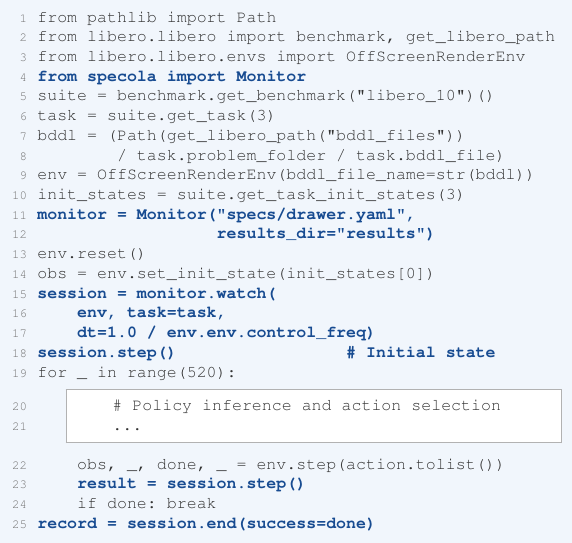}
\caption{An adapted LIBERO rollout for OpenVLA-OFT. The \toolname{} import and four calls are marked in blue; the box represents policy code.}
\label{lst:loop}
\end{codelisting}

\noindent \textbf{Standalone evaluation:} \toolname{} also provides command-line tools for running and inspecting monitors without writing an evaluation loop (Table~\ref{tab:cli}). The \texttt{run} command executes a single episode, either by replaying a recorded demonstration or applying actions in the simulator, while streaming robustness values and optionally displaying them in a browser dashboard. The resulting robustness traces are saved for later analysis. The \texttt{report} command summarizes these saved runs with statistics and plots grouped by task and policy, while \texttt{view} provides a browser-based replay of an individual run, including its video and robustness trace, without requiring the simulator.

\begin{table}[!htb]
\caption{Main command-line arguments and outputs.} %
\label{tab:cli}
\centering
\footnotesize
\setlength{\tabcolsep}{4pt}
\renewcommand{\arraystretch}{1.05}
\begin{tabularx}{\columnwidth}{@{}p{0.67\columnwidth} L@{}}
\toprule
\textbf{Command and arguments} & \textbf{Output} \\
\midrule
\textbf{\texttt{discover}}\newline
\texttt{[-{}-ui UI] [-{}-web [PORT]]} &
Entities and signals \\
\addlinespace[2pt]
\textbf{\texttt{write-specs}}\newline
\texttt{[-{}-output FILE] [-{}-ui UI]} &
Validated YAML specifications \\
\addlinespace[2pt]
\textbf{\texttt{check}} \texttt{FILE} &
Resolved signals and bindings \\
\addlinespace[2pt]
\textbf{\texttt{run}} \texttt{-{}-specs FILE}\newline
\texttt{[-{}-mode MODE] [-{}-demo FILE]}\newline
\texttt{[-{}-steps N] [-{}-web [PORT]]} &
Per-step robustness; optional dashboard \\
\addlinespace[2pt]
\textbf{\texttt{report}} \texttt{RESULTS\_DIR}\newline
\texttt{[-{}-by COLUMNS] [-{}-csv FILE]}\newline
\texttt{[-{}-parquet FILE]} &
Saved-run summary \\
\addlinespace[2pt]
\textbf{\texttt{view}} \texttt{RECORDING\_DIR [-{}-port PORT]} &
Recording in a browser \\
\bottomrule
\end{tabularx}
\par\smallskip
\begin{minipage}{\columnwidth}
\footnotesize
Commands are in bold, optional arguments are in brackets. The first four commands also accept
\texttt{[-{}-sim SIM]} and \texttt{[-{}-task TASK]}.
\texttt{SIM}, \texttt{TASK}, and \texttt{UI} select the adapter, task,
and interface. \texttt{FILE} and \texttt{DIR} values are paths;
\texttt{N} and \texttt{PORT} are integers.
\texttt{MODE} is \texttt{random}, \texttt{hold}, or \texttt{replay};
replay requires \texttt{-{}-demo FILE}.
\texttt{COLUMNS} lists comma-separated metadata columns.
\end{minipage}
\end{table}

\subsection{Adding New Simulators}
\label{sec:adapter}
An adapter is a Python class that turns a simulator's state into
\toolname{} observations. It must provide four things: the monitoring
period \inlinecode{dt}, the number of spatial dimensions, a reset hook
that returns the first observation of an episode, and an observation hook
that returns the current one. An observation holds a step index, a time
in seconds, and the entities in the scene. Each entity has a class, an
identifier that stays fixed for the episode, an optional position and
velocity, and named scalar properties. 
Rendering, camera, and scene-geometry hooks are optional. An adapter
registers under a \inlinecode{-{}-sim} name, either from its own package
or inside the repository. For MuJoCo-based simulators, a shared MuJoCo
adapter already does most of the work; a new adapter only supplies rules
for classifying entities. The tool documentation provides the template and implementation details;
our coding-agent skill guides adapter writing. %

\section{Case Studies}
\label{sec:case-studies}

The following case studies demonstrate how \toolname{} can be used to
provide insight and ensure correct evaluation of policy performance. We show how the robustness score can help detect \textit{false failures} that may result in incorrect success-rate-based policy comparisons (Sec.~\ref{sec:falseFail}), identify \textit{failure modes} (Sec.~\ref{sec:failMod}),  detect \textit{unstable successes} that might become failures (Sec.~\ref{sec:case-goal-hold}), and evaluate behaviors beyond success criteria (Sec.~\ref{sec:beyond}). %

\subsection{Experimental Setup}
\label{sec:case-setup}

We evaluated policy rollouts in LIBERO and RoboDojo. For
LIBERO~\cite{liu2023libero}, we evaluated rollouts of
OpenVLA~\cite{kim2024openvla}, OpenVLA-OFT~\cite{kim2025openvlaoft},
VLA-Adapter~\cite{wang2026vlaadapter}, and \pizero{}~\cite{black2025pi05}
across 40 tasks from the Spatial, Object, Goal, and Long suites.
Each policy was run 50 times; all policies were run with the same set of initial conditions. %
For RoboDojo~\cite{chen2026robodojo}, we used
XPolicyLab~\cite{community2026xpolicylab} to run
MolmoAct2~\cite{fang2026molmoact2}, \pizero{}~\cite{black2025pi05}, and
StarVLA PI-v3~\cite{community2026starvla} on stacking three bowls.
Each policy was evaluated on 25 initial conditions.

For all tasks, we wrote goal specifications that mirror the benchmark success criteria. Each goal is a conjunction of the predicates the benchmark checks; we also monitor each predicate on its own to see which failed (Section~\ref{sec:failMod}). %

We used an Intel Core Ultra~7 265K CPU, 64\,GB of RAM, and an NVIDIA GeForce RTX~5090 GPU with 32\,GB of memory to run the simulators and monitors.

\noindent \textbf{Monitoring overhead:} To examine the compute time and memory overhead induced by running the monitor during execution, we use 36 tasks in LIBERO and measure the time and memory for 3 runs per task; one run without monitoring, one with state extraction only, and one with full
specification evaluation.   
In all 36 tasks, the three runs produced byte-identical actions, simulator states, termination flags, and outcomes.
Full monitoring lengthened the control loop time by $17.6\% \pm 7.8$\%. %
Per step, state extraction cost
$3.3 \pm 1.1$\,ms, specification evaluation $0.49 \pm 0.10$\,ms, and
bookkeeping $0.04 \pm 0.01$\,ms, against $15.7 \pm 0.8$\,ms of policy
inference and $7.2 \pm 1.3$\,ms of simulation. Process memory grew by
$34 \pm 29$\,MB.

\subsection{Failures that should be successes}\label{sec:falseFail}
To obtain a success/fail decision, benchmarks implement, internally, functions that compare privileged state (object location, end effector pose, etc) to thresholds. These thresholds are typically hand designed, and while in general good at separating success from failure, they are not perfect. The result is that success rate can be skewed by these (internal) thresholds, leading us to conclusions that might not be accurate. The robustness scores that \toolname{} provides can help identify these situations.

\begin{figure}[!htb]
\centering
\includegraphics[width=\columnwidth]{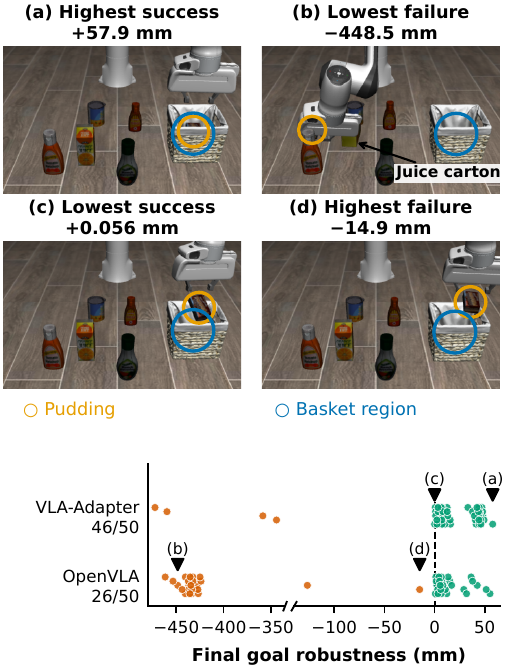}
\caption{VLA-Adapter and OpenVLA on the pudding-in-basket task (LIBERO
Object task 8, 50 rollouts each). Panels (a) and~(c) show VLA-Adapter's highest and lowest successes, panels (b) and~(d) show OpenVLA's lowest and highest failures. The plot shows the goal robustness for all 50
rollouts of each policy with the illustrated trials marked.}%
\label{fig:case-pudding-extremes}
\end{figure}

\noindent \textbf{Case 1:} Figure~\ref{fig:case-pudding-extremes} depicts the LIBERO Object task ``pick up the chocolate pudding and place it in the basket''. At the bottom it graphs the robustness of the goal specification for each of the 50 rollouts for two of the policies we tested; a positive number is a success, a negative number is a failure. Panels (c) and (d) depict a success and a failure, respectively. Semantically, they seem to be the same; slightly in the basket but not in its middle.
Panel (c) counts as a success by 0.056\,mm and panel (d) as a failure by 14.9\,mm; the two rollouts differ by less than the pudding's width yet receive opposite labels. Panels (a) and (b) show that the same policies also produce outcomes the label does describe well: a placement well inside the basket at $+57.9$\,mm, and a rollout that placed the juice carton in the basket instead of the pudding, at $-448.5$\,mm. The success rates, 46 of 50 for VLA-Adapter and 26 of 50 for OpenVLA, treat (c) and (d) as the same kind of difference as (a) and (b). Robustness separates the two, and shows how the success rate would change under a different threshold. %

\begin{figure}[!htb]
\centering
\includegraphics[width=\columnwidth]{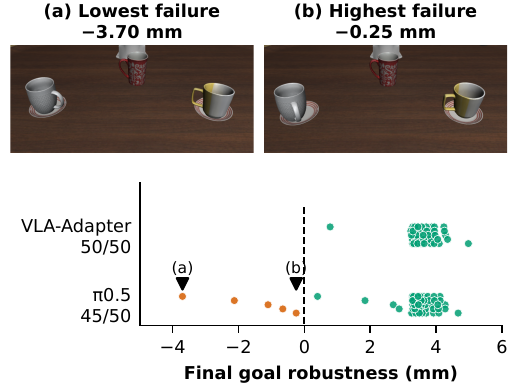}
\caption{VLA-Adapter and \pizero{} on the two-mug task (LIBERO-10 task 4, 50 rollouts each). Panels (a) and~(b) show \pizero{}'s failures with the lowest and highest goal robustness. The plot shows all 50 rollouts of each policy with the illustrated trials marked. %
}%
\label{fig:case-mug-policies}
\end{figure}

\noindent \textbf{Case 2:} Figure~\ref{fig:case-mug-policies} depicts the LIBERO-10 task ``put the white mug on the left plate and put the yellow and white mug on the right plate''. The benchmark counts a rollout as a success when each mug touches its plate with its center above the plate's center and less than 30\,mm from it horizontally. VLA-Adapter succeeded in 50 of its 50 rollouts and \pizero{} in 45. At the bottom, the figure graphs the robustness of the goal specification for each rollout of the two policies. All five \pizero{} failures lie within 4\,mm of the threshold. %
Panels (a) and (b) depict the failures with the lowest and highest goal robustness. In these and the three other \pizero{} failures, both mugs rest on their plates, so judged against the natural-language task these rollouts are successes. The robustness values led us to inspect them; counting them as successes raises \pizero{} from 90\% to 100\%, indistinguishable from VLA-Adapter, in contrast to what the benchmark outcome would suggest. %

\subsection{Failure modes}\label{sec:failMod}
A binary failure label does not capture by how much or on which requirement the rollout failed. %
Because each goal specification is the conjunction of the benchmark's predicates, the monitor can report the robustness of every predicate alongside the goal, and these scores can provide insight into policy failure modes.

\begin{figure}[!htb]
\centering
\includegraphics[width=\columnwidth]{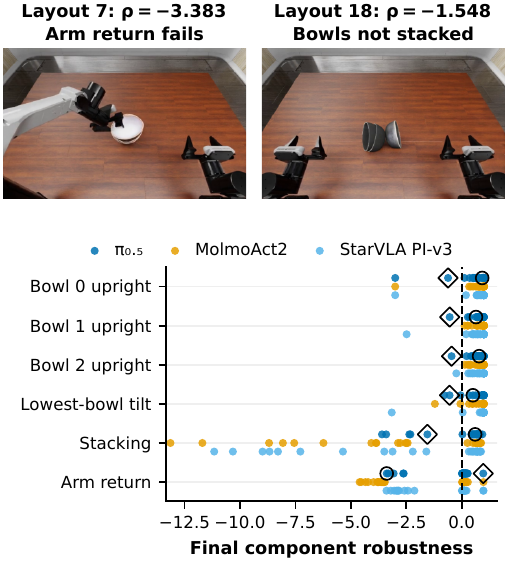}
\caption{Robustness for all 25 bowl-stacking rollouts
per policy, including successes and failures. Dot colors distinguish
\pizero{}, MolmoAct2, and StarVLA PI-v3. Images above show \pizero{} layouts~7 and~18, marked in the plot by a ring and a diamond.}%
\label{fig:case-bowl-opposite-failures}
\end{figure}

\noindent \textbf{Case 3:} Figure~\ref{fig:case-bowl-opposite-failures} depicts RoboDojo's bowl-stacking task. The benchmark scores a rollout 15 if any two bowls are stacked upright and the gripper is open, 100 if all three bowls are stacked upright, the bottom bowl rests stably on the table, and the robot returns to its origin, and 0 otherwise. The goal requires each bowl's tilt be below $45^\circ$ and the lowest bowl's tilt below $7^\circ$. After sorting the bowls by height, adjacent centers must be at most 4\,cm apart horizontally and at least 5\,mm apart vertically (the \inlinecode{Stacking} predicate), and both arms must return to within 15\,cm and $20^\circ$ of their initial end-effector poses. \pizero{} succeeded on 19 of 25 layouts, MolmoAct2 on 11, and StarVLA PI-v3 on 13, and the benchmark gave all six \pizero{} failures the same process score of 15. The plot in Figure~\ref{fig:case-bowl-opposite-failures} graphs the final robustness of each of the six requirements for every rollout of the three policies. Two of \pizero{}'s failures failed in opposite ways. In layout~7 (left image), the bowls were stacked and upright but the arms did not return; in layout~18 (right image), the arms returned but the bowls had tipped over. The other four failures missed both. Looking at the plot, we can see that all policies were almost 100\% successful in keeping bowls upright, but \pizero{} was the most robust at stacking them. %

\subsection{Unstable successes}\label{sec:case-goal-hold}
The benchmarks typically check for a goal state. %
A rollout can satisfy the goal at a specific timestep and violate it immediately after. STL can capture temporal %
specifications, allowing analysis of the temporal robustness of the policy. %

\begin{figure}[!htb]
\centering
\includegraphics[width=\columnwidth]{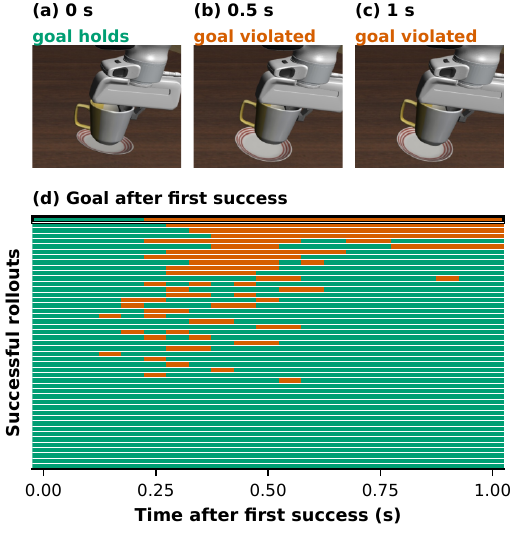}
\caption{OpenVLA-OFT on the two-mug task after the first success.
Panels~(a)--(c) show trial~37 at 0, 0.5, and 1\,s. Panel~(d) shows the
benchmark check at each control step for all 47 successes, one row per
rollout; green marks the goal holding, orange a violation, and the outlined
row is trial~37.}%
\label{fig:case-goal-hold}
\end{figure}

\noindent \textbf{Case 4:} We reran OpenVLA-OFT's 47 successful two-mug rollouts, letting the policy act for 20 more control steps (1\,s) after the success, and monitored \inlinecode{historically[0,1s](goal)}, which is true when the goal has held throughout the past second. Each rerun reproduced the recorded simulator states exactly up to the first success. The goal held for the full second in only 16 of the 47 rollouts (Fig.~\ref{fig:case-goal-hold}(d)). For example, in trial~37 (panels (a)--(c)), the mug touched the plate at the first success while still held by the gripper; the robot then lifted the cup again, by 26.5\,mm. %
 The benchmark counts the 47 rollouts as successes; however, some of the success determinations are premature and it is possible that some rollouts should be labeled as failures. %

\subsection{Beyond ``success''}\label{sec:beyond}
The cases up until now focused on the specification that captures the success criteria. Another advantage of having a tool to write and evaluate specifications is the ability to evaluate other aspects of the robot behavior.

\begin{figure}[!htb]
\centering
\includegraphics[width=\columnwidth]{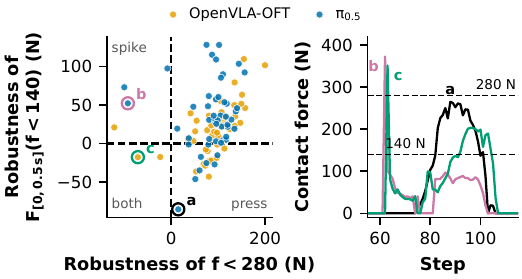}
\caption{Left: robustness of the contact specification for successful stove rollouts, $f<280$ on the horizontal axis and $\Fop{[0,\,0.5\,\mathrm{s}]}(f<140)$ on the vertical; points left of or below the dashed lines violate that part. Right: the contact force on the stove button over steps 55 to 115 in the circled rollouts: (a) a sustained press, (b) a spike above 280\,N, and (c) both. %
}%
\label{fig:case-contacts}
\end{figure}

\noindent \textbf{Case 5:} Figure~\ref{fig:case-contacts} shows the LIBERO-10 task ``turn on the stove and put the moka pot on it.'' OpenVLA-OFT succeeded in 49 of 50 rollouts and \pizero{} in all 50. We monitored the contact requirement in Listing~\ref{lst:spec}, which limits the force on the robot to 140\,N, following ISO/TS~15066~\cite{iso15066}. Despite their high success rates, 42 of OpenVLA-OFT's 49 successes and 47 of \pizero{}'s 50 exceeded this limit, in each case while a finger pressed the stove button. These violations differed in duration and magnitude. Most were brief rises; others were sustained presses, such as rollout (a). %
The original requirement does not distinguish the two. We therefore introduced a specification with two conditions: the contact force must remain below 280\,N, and any force above 140\,N must return below it within 0.5\,s: $(f < 280) \wedge \Fop{[0,\,0.5\,\mathrm{s}]}(f < 140)$. %
It was violated by 15 of OpenVLA-OFT's 49 successful rollouts and 12 of \pizero{}'s 50: 14 and 9 by sustained presses, as in (a), and the rest by single-step spikes above 280\,N, as in (b). Two of OpenVLA-OFT's sustained presses also spiked above 280\,N, as in (c). Neither success rate nor the original threshold captures these distinctions.

\section{Discussion}

We presented \toolname{}, a tool for evaluating robot behavior in simulation. Writing and evaluating signal temporal logic specifications over the robot state and privileged simulator signals, \toolname{} allows researchers to gain a more nuanced understanding of robot behavior, compared to using the typical success rate. Furthermore, \toolname{} allows us to also reason about the validity of the success rate, as we have shown in Sec.~\ref{sec:falseFail} and \ref{sec:case-goal-hold}.

Additional insights that can be gained using \toolname{} include assessing the robustness of a successful policy and evaluating requirements that are not represented in the success criteria (as illustrated in Sec.~\ref{sec:beyond}). For the former, if there are several policies with the same success rate, looking at the robustness score can provide details regarding \textit{how} the policy behaves. For the later, we can add specifications about any signal in the simulator; these specifications can be complex and include timing and conditions such as constraints on gripper width and timing or detecting periods when the robot freezes. 

In addition to bulk examination of policy performance, the robustness score allows researchers to pinpoint interesting rollouts to visualize. It is common practice to run policies on hundreds of initial conditions; it is not practical to visualize and view all the rollouts. As we have shown in the case studies, visualizing a few rollouts, chosen based on their robustness, allows us to gain further insight. %

\toolname{} currently monitors simulation using privileged state, with one observation per action at a fixed control period. On physical robots, however, sensors produce observations at different times and with different delays. Furthermore, the inherent uncertainty in physical systems requires reasoning about perception failures in the monitoring pipeline. Our future plans include extending \toolname{} to evaluate physical robots by addressing these challenges.

\bibliographystyle{IEEEtran}
\bibliography{bstctl,references}

\end{document}